\pdfoutput=1

\documentclass[11pt]{article}
\usepackage[]{emnlp2021}

\usepackage{times}
\usepackage{latexsym}
\usepackage[T1]{fontenc}
\usepackage{pgfplots}
\usepackage{balance}
\usepackage[all]{nowidow}
\usepackage{newtxtext}
\usepackage{newtxmath}
\usepackage{subcaption}

\usepackage{graphicx} 

\usepackage[]{todonotes}

\DeclareMathAlphabet{\pazocal}{OMS}{zplm}{m}{n}
\newcommand{\unif}{\pazocal{U}}

\usepackage{microtype}

\DeclareMathOperator*{\LCS}{\mathrm{LCS}}
\DeclareMathOperator*{\LD}{\mathrm{LD}}

\usepackage{silence}
\usepackage{mathtools}
\usepackage{amsthm}
\usepackage{bm}
\usepackage{color}
\usepackage{soul}           
\usepackage{url}
\usepackage{booktabs}       
\usepackage{nicefrac}       
\usepackage{enumerate}
\usepackage{graphicx}
\DeclareGraphicsExtensions{.pdf,.png,.jpg}
\graphicspath{{figures/}}
\usepackage{natbib}
\usepackage{comment}
\usepackage{hyperref}
\usepackage{centernot}     

\usepackage[norelsize, algo2e, ruled, lined, linesnumbered]{algorithm2e}
\SetAlFnt{\small}

\usepackage{multicol,lipsum}
\usepackage{cuted, nccmath}
\usepackage{multirow}
\usepackage{bm}

\usepackage{amsfonts}       

\newcommand{\parg}{\makebox[1ex]{$\mathbf{\cdot}$}}                              
\newcommand{\mquad}{\kern-1em}													

\newcommand{\InNorm}[1]{{\left\vert\kern-0.2ex\left\vert\kern-0.2ex\left\vert #1 
    \right\vert\kern-0.2ex\right\vert\kern-0.2ex\right\vert}}                    

\newcommand{\InNormII}[1]{{\left\vert\kern-0.2ex\left\vert\kern-0.2ex\left\vert #1 
    \right\vert\kern-0.2ex\right\vert\kern-0.2ex\right\vert}_2}                    

\newcommand{\InNormInfty}[1]{{\left\vert\kern-0.2ex\left\vert\kern-0.2ex\left\vert #1 
    \right\vert\kern-0.2ex\right\vert\kern-0.2ex\right\vert}_{\infty}}           

\newcommand{\Abs}[1]{\ensuremath{\left \lvert #1 \right \rvert}}                 
\DeclarePairedDelimiterX{\Inner}[2]{\langle}{\rangle}{#1, #2}                    

\definecolor{gray}{rgb}{0.7,0.7,0.7}

\theoremstyle{definition}

\title{XLEnt: Mining a Large Cross-lingual Entity Dataset with Lexical-Semantic-Phonetic Word Alignment}

\author{Ahmed El-Kishky$^1$ \enskip
Adithya Renduchintala$^2$ \enskip
James Cross$^2$ \\
\textbf{Francisco Guzm\'an$^2$ \enskip
Philipp Koehn$^3$} \\
$^1$Twitter Cortex, $^2$ Facebook AI \enskip $^3$Johns Hopkins University \\
{\tt aelkishky@twitter.com, \{adirendu,jcross,fguzman\}@fb.com, phi@jhu.edu} \\
}

\date{}

\begin{document}
\maketitle
\begin{abstract}
Cross-lingual named-entity lexica are an important resource to multilingual NLP tasks such as machine translation and cross-lingual wikification. While knowledge bases contain a large number of entities in high-resource languages such as English and French, corresponding entities for lower-resource languages are often missing. To address this, we propose Lexical-Semantic-Phonetic Align (LSP-Align), a technique to automatically mine cross-lingual entity lexica from mined web data. We demonstrate LSP-Align outperforms baselines at extracting cross-lingual entity pairs and mine 164 million entity pairs from 120 different languages aligned with English. We release these cross-lingual entity pairs along with the massively multilingual tagged named entity corpus as a resource to the NLP community.
\end{abstract}

\section{Introduction}

Named entities are references in natural text to real-world objects such as persons, locations, or organizations that can be denoted with a proper name. Recognizing and handling these named entities in many languages is a difficult, yet crucial, step to language-agnostic text understanding and multilingual natural language processing (NLP)~\cite{sekine2009named}.

As such, cross-lingual named entity lexica can be invaluable resources towards making tasks such as entity linking, named entity recognition~\cite{ren2016automatic1,ren2016automatic2}, and information and knowledge base construction~\cite{tao2014newsnetexplorer} inherently multilingual. However, the coverage of many such multilingual entity lexica (e.g., Wikipedia titles) is less complete for lower-resource languages and approaches to automatically generate them under-perform due to the poor performance of low-resource taggers~\cite{feng2018improving,cotterell2017low}.

\begin{figure}
    \centering
    \includegraphics[width=8cm]{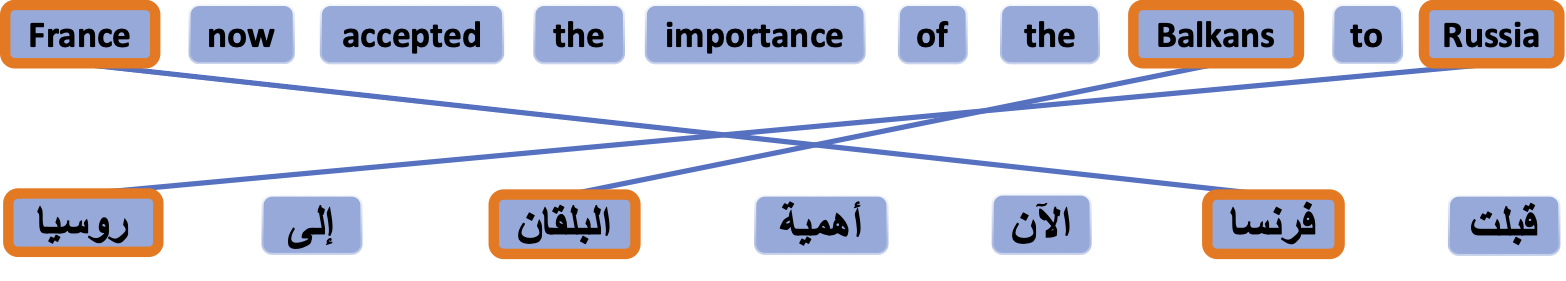}
    \caption{Identify entity pairs by projecting English entities onto lower-resource languages via word-alignment. }
    \label{fig:example}
\end{figure}

To perform low-resource NER, previous efforts have applied word alignment techniques to project available labels to other languages. \citet{kim2010cross} applies heuristic approaches with alignment correction using an alignment dictionary of entity mentions. \citet{das2011unsupervised} introduced a novel label propagation technique that creates a tag lexicon for the target language, while \citet{wang2014cross} instead projected model expectation rather than labels thus transferring word boundary uncertainty. Additional work jointly performs word alignment while training bilingual name tagging~\cite{wang2013joint}; however this method assumes the availability of named entity taggers in both languages. Other methods have leveraged bilingual embeddings for projection~\cite{ni2017weakly, xie2018neural}.

In this work, we propose using named-entity projection to automatically curate a large cross-lingual entity lexicon for many language pairs. As shown Figure~\ref{fig:example}, we construct this resource by performing NER in a higher-resource language, then projecting the entities onto text in a lower-resource language using word-alignment models.

Our main contribution is the construction and release of a large web-mined cross-lingual entity dataset that will be beneficial to the NLP community. Our proposed alignment model, LSP-Align, principally combines the lexical, semantic, and phonetic signals to extract higher-quality cross-lingual entity pairs as verified on a ground-truth entity pair set. With LSP-Align, we mined over \textit{164M} distinct cross-lingual entity pairs spanning \textit{120} language pairs and freely release the XLEnt dataset\footnote{\url{http://data.statmt.org/xlent/}}\footnote{\url{https://opus.nlpl.eu/XLEnt-v1.1.php}} in hope it spurs further work in cross-lingual NLP.

\section{Preliminaries}

We formally define an entity collection as a collection of extracted text spans tied to named entity mentions. We denote these named entity mentions as $M =\{ne_i\}_{i=1}^n$ , where $ne_i$ is the $i_{th}$ named entity in the mention collection $M$ and $n$ is the size of $M$. 

Cross-lingual entity lexicon creation seeks to create two entity collections $M_1$ and $M_2$ in a source and target language respectively. These two collections should be generated such that for each entity mention in $ne_i \in M_1$ in the source language, there is a corresponding named entity $ne_j \in M_2$ in the target language such that $ne_i$ and $ne_j$ refer to the same named entity in their respective language.


\section{Mining Cross-lingual Entities}

We introduce our approach to automatically extract cross-lingual entity pairs from large mined corpora.

\subsection{High-Resource NER}
We begin with large collections of comparable bitexts mined from large multilingual web corpora~\cite{el2020searching}. In particular, we select three mined web corpora 1) CCAligned~\cite{el2020massive}, 2) WikiMatrix~\cite{schwenk2019wikimatrix}, and 3) CCMatrix~\cite{schwenk2019ccmatrix}) due to the wide diversity of language pairs available in these mined corpora. We select language pairs of the form English-Target and tag each English sentence with named entity tags~\cite{ramshaw1999text} using a pretrained NER tagger provided in the Stanza NLP toolkit\footnote{\url{https://stanfordnlp.github.io/stanza/}}~\cite{qi2020stanza}. This NER model adopts a contextualized string representation-based tagger proposed by~\citet{akbik2018contextual} and utilizes a forward and backward character-level LSTM language model. At tagging time, the representation at the end of each word position from both language models with word embeddings is fed into a standard Bi-LSTM sequence tagger with a conditional-random-field decoder.

\subsection{Entity Projection via Word Alignment}
\label{subsec:signals}
We introduce three approaches for projecting entities and LSP-Align which combines all three.

\subsubsection{Lexical Alignment}
To perform word alignment using lexical-cooccurences, we apply FastAlign~\cite{dyer2013simple}, a fast loglinear re-parameterization of IBM Model 2~\cite{brown1993mathematics} and symmetrize alignments using the grow-diagonal-final-and (GDFA) heuristic. 

FastAlign performs unsupervised word alignment over the full collection of mined bitexts using an expectation maximization based algorithm. While FastAlign is state-of-the-art in word alignment, due to its reliance on lexical co-occurences, it may misalign low-frequency entities.

\subsubsection{Semantic Alignment}
We leverage multilingual representations (embeddings) from the LASER  toolkit~\cite{artetxe2019massively} to align words that are semantically close. We propose a simple greedy word alignment algorithm guided by a distance function between words:

\begin{equation}
sem(w_s,w_t) = 1 - \frac{\mathbf{v}_{s} \cdot \mathbf{v}_{t}}{||\mathbf{v}_{s}||~ ||\mathbf{v}_{t}||}
\label{eq:cosine}
\end{equation}

\begin{algorithm2e}
\caption{Distance Word Alignment}
\label{alg:greedy}
\Indm
      \KwIn{$P = \{(w_s, w_t)~|~w_s \in S_s, w_t \in S_t\}$} 
      \KwOut{$P'= \{(w_{s,i},w_{t,i}), . . .\} \subset P$}
\Indp

      \BlankLine
    $word{-}pairs$ $\gets \{(p, dist(p)) \text{ for } p \in P\}$ \\
	$sorted$ $\gets sort(word{-}pairs) \text{ in ascending order } $\\
	aligned, $~S_s,~S_t$ $\gets \varnothing,~\varnothing,~\varnothing$\\
	$free \gets ||S_s| - |S_t||$ \\
	\For{$w_s, w_t \in$ sorted}{
		\uIf{$w_s \notin S_s \land w_t \notin S_t$}  
		{
		   $aligned \gets aligned \cup \{(w_s, w_t)\}$ \\
		   $S_s \gets S_s \cup w_s$ \\
		   $S_t \gets S_t \cup w_t$ \\
		}
	    \uElseIf{$free > 0 \land |S_s| < |S_t| \land w_s \in S_s$}  
		{
		   $aligned \gets aligned \cup \{(w_s, w_t)\}$ \\
		   $S_t \gets S_t \cup w_t$ \\
		   $free \gets free{-}1$\\
		}
		\uElseIf{$free > 0 \land |S_s| > |S_t| \land w_t \in S_t$}  
		{
		   $aligned \gets aligned \cup \{(w_s, w_t)\}$ \\
		   $S_w \gets S_w \cup w_w$ \\
		   $free \gets free{-}1$\\
		}	
	}	
	  \textbf{return} aligned
\end{algorithm2e}

where Equation~\ref{eq:cosine} shows that the semantic distances between a source word ($w_s$) and target word ($w_t$) is simply $1$ minus the cosine similarity between $v_s$ and $v_t$, the LASER vector representations of $w_s$ and $w_t$ respectively. As shown in Algorithm~\ref{alg:greedy}, we take each source-target sentence pair and perform alignment between their tokens guided by the semantic distances between words. Of course, as source and target sentences, may be of different sizes, tokens in the shorter sentence may be aligned with multiple target tokens. Unlike lexical alignment with FastAlign, our distance-based alignment is deterministic and only needs a single pass through the bitexts.

\subsubsection{Phonetic Alignment}
Recognizing that in many cases, phonetic transliterations are the avenue by which proper names travel between languages, we propose using phonetic signals to perform alignment and match named entities.

To align words based on their phonetic similarity, we leverage the distances between their transliterations and align words between the source and target that are ``close" in this phonetic space. We adopt an unsupervised transliteration system developed by~\citep{chen2016false} to transliterate between source and target languages and utilize Levenshtein distance (aka edit distance)~\cite{wagner1974string} to calculate distances between transliterated words:

\begin{small}
\begin{equation}
\label{eq:phonetic}
\arraycolsep=1.4pt\def\arraystretch{2.2}
    phon(w_s,w_t) = \min \left\{\begin{array}{l}
        \LD(T_{w_s}, w_t)) / {\max(|T_{w_s}|, |w_t|)} \\
        \LD(w_s, T_{w_t})) / {\max(|w_s|, |T_{w_t}|)}\\
        \LD(w_s, w_t) / {\max(|w_s|, |w_t|)}
        
        \end{array}\right\} 
\end{equation}
\end{small}

 where $\ LD(\parg, \parg)$ is the \emph{Levenshtein distance} between two strings and $T_a$ is the transliteration of word $a$ into word $b$'s language. Equation~\ref{eq:phonetic} selects the minimum normalized distance between a source transliteration, target transliteration, and no transliteration to guide Algorithm~\ref{alg:greedy} for a greedy word alignment. Once again, only a single pass over the data is required for alignment.

\subsubsection{Estimating Translation Probabilities}
Leveraging lexical alignment (i.e, FastAlign) alongside semantic and phoentic alignment yields three potential word alignments for a bitext collection. For alignment method $k$, we can iterate through the alignments and compute the counts of source-to-target $(s, t)$ word pairings; we denote this count $cnt(s,t)$. We can estimate the maximum likelihood translation probability from $s$ to $t$ given by alignment method $k$ as follows:
\begin{equation}
    \theta_{k,s,t} = \frac{cnt(s,t)}{\sum_{t'} cnt(s, t')}
    \label{eq:maximization}
\end{equation}

Using Equation~\ref{eq:maximization}, we can compute the translation probabilities for lexical, semantic, and phonetic alignments which we use in our LSP-Align model.

\subsection{LSP Named-entity Projection}
 We describe LSP-Align, which combines the three alignment signals for better entity-pair mining. 

\begin{algorithm2e}
\caption{LSP-Align Generative Model}
\label{alg:generative}
\Indm
      \KwIn{$S=\{s_1 \ldots s_m\}$ \tcp{source sentence}}
      \KwOut{$T=\{t_1 \ldots t_n$\} \tcp{translated sentence}}
\Indp

      \BlankLine
    let $\theta_k$: $k \in\{1, 2, 3\}$ be the translation distributions \tcp{1=lexical, 2=semantic, 3=phonetic}
    
    draw length $n$ for translation $T$ using $|S|=m$ \\

	\For{each $j \in 1\ldots n$}{
		draw $a_j\in \{0,1,\ldots, m\} \sim \unif(0,m)$ \\
		draw $k_j \sim \unif(1,3)$ \\
        draw $t_j \sim \theta_{k_j,s_{a_j},t_j}$ 
	}	
	\textbf{return} $T$
\end{algorithm2e}
\begin{figure}
    \centering
    \includegraphics[width=5cm]{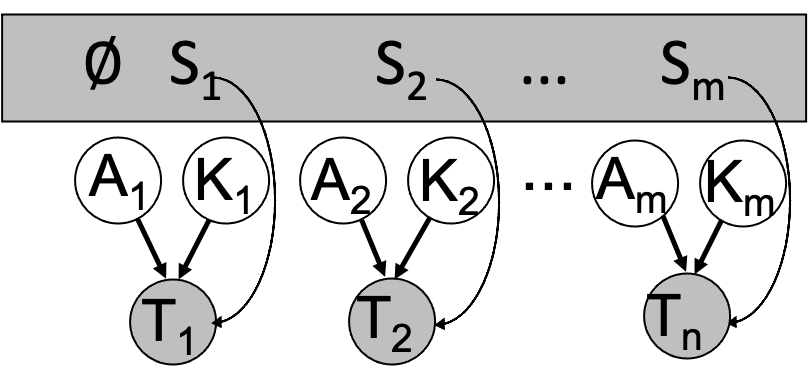}
    \caption{S and T are source/target sentences; target words are drawn from a distribution determined by (1) alignment, (2) source word, and (3) translation method }
    \label{fig:plate}
\end{figure}
As described in Algorithm~\ref{alg:generative}, the generative process takes in a source sentence $S$ and translates this sentence into the target sentence by drawing an alignment variable and translation mechanism (lexical, semantic, or phonetic) for each position in the target sentence and drawing a translated word from the corresponding translation distribution. 


\begin{table*}[t]
\small
    \centering
    \begin{tabular}{l  l  l l c c c c}
        \toprule
       \textbf{Resource} & \textbf{Language} & \textbf{Num Bitexts} & \textbf{Distinct Ents} &  \textbf{Lexical} & \textbf{Semantic} & \textbf{Phonetic} & \textbf{LSP-Align}   \\\midrule
        
       \multirow{2}{*}{{\bf High} } &  Russian  & 3.2M & 40.4K & 0.84 & 0.81  & 0.83  &  \textbf{0.86}\\
       
       & Chinese    & 5.2M & 28.4K  &  \textbf{0.85} & 0.78   & 0.73  &  \textbf{0.85}\\
       
       & Turkish    & 2.5M & 27.4K  &  0.88 & 0.89  & 0.87  &  \textbf{0.90}
       
       \\\midrule

       \multirow{2}{*}{{\bf Mid}}  & Arabic  & 4.9M & 26.4K & \textbf{0.88}  & 0.80 & 0.81  &  \textbf{0.88}\\
       
        & Hindi  & 1.2M & 7.60K & 0.89 & 0.73  & 0.87  &  \textbf{0.90}\\
       
        & Romanian   & 2.1M & 26.2K  & 0.93 & \textbf{0.94}  & 0.92  &  \textbf{0.94}
        
        \\\midrule

        \multirow{2}{*}{{\bf Low }} & Estonian  & 1.3M & 15.2K & 0.87   & \textbf{0.89}  & 0.87  &  \textbf{0.89}\\
        
        & Armenian  & 52K & 2.30K  & 0.78 & 0.44   & \textbf{0.83}  &  0.81 \\
        
        & Tamil   & 45K & 2.50K  & 0.67 & 0.50   & 0.71  &  \textbf{0.72}\\
        
        \midrule
        & \textbf{Avg} & - & - & 0.84  & 0.75 & 0.83  &  \textbf{0.86}
        \\\bottomrule
        
    \end{tabular}
    \caption{Fuzzy-F1 scores of mined cross-lingual entity pairs evaluated against gold-standard pairs.}
    \label{tab:gold_extraction}
\end{table*}


The graphical model for LSP-Align depicted in Figure~\ref{fig:plate}, is similar to IBM-1~\cite{brown1993mathematics}. The main difference is that, in addition to latent alignment variables $A$, we introduce latent translation mechanisms $K$. The translation distribution $\theta_{K,s}$ is chosen based on the latent alignment and mechanism variables. As we demonstrate in Equation~\ref{eq:maximization}, we can leverage the alignments for each alignment signal to estimate $\theta_{K,s}$ for each translation distribution. Using these estimated distributions in our model, we can infer the alignment variables as follows:

\begin{equation} \label{eq:alignment_prob}
\small
\begin{split}
 P(a_j = i | S, T, \theta)  & = \sum_{k_j{=}1}^3 P(a_j{=}i | S, T, k_j, \theta) \cdot P(k_j) \\
                            & = \sum_{k_j{=}1}^3 \theta_{k_j,s_i,t_j}\cdot \frac{1}{3} \\
\end{split}
\end{equation}

where we assign the most probable alignment variable to each target word after marginalizing over the latent translation mechanisms (lexical, semantic, phonetic), which, for simplicity, we give equal probability.

\section{Experiments \& Results}
\paragraph{Datasets}
We utilize a gold standard evaluation lexicon created by~\cite{pan2017cross} that leverages eight named parallel entity corpora\footnote{Chinese-English Wikinames, Geonames, JRC names, LORELEI LRLP, NEWS 2015 task, Wikipedia names, Wikipedia places, and Wikipedia titles}. We select nine languages from a diverse set of resource availability, language families, and scripts for evaluation.

\paragraph{Evaluation Protocol}
We evaluated the performance of the methods using the commonly used fuzzy-f1 score~\cite{Tsai2018LearningBN} which is defined as the harmonic mean of the fuzzy precision and fuzzy recall scores. This metric is based on the longest common subsequence between a gold and mined entity, and has been used for several years in the NEWS transliteration workshops~\cite{li2009report,banchs2015report}. The fuzzy precision and recall between a predicted string $p$ and the correct string $t$ is computed as follows:

\begin{align*}
    \mathrm{fuzzy{-}precision}(p,t) &= \nicefrac{\Abs{\LCS(p, t)}}{\Abs{p}}, \\
    \mathrm{fuzzy{-}recall}(p,t) &= \nicefrac{\Abs{\LCS(p, t)}}{\Abs{t}},
\end{align*}

where $\ LCS(\parg, \parg)$ is the \emph{longest common subsequence} between two strings.

\subsection{Cross-lingual Entity Extraction}
\label{sec:extraction}
We take a small sample of parallel sentences for each language, mine entity pairs using each projection technique, and compute Fuzzy-F1 using the gold-standard as a reference. As seen in Table~\ref{tab:gold_extraction}, while lexical alignment outperforms semantic alignment, it displays similar performance to phonetic with phonetic performing better on low-resource languages and lexical performing better on high-resource. However, LSP-Align outperforms or matches lexical alignment consistently showing that using all signals yields superior NE projection.

\begin{figure}
  \subcaptionbox{low-frequency \label{low-freq}}{{%
      \includegraphics[width=0.23\textwidth]{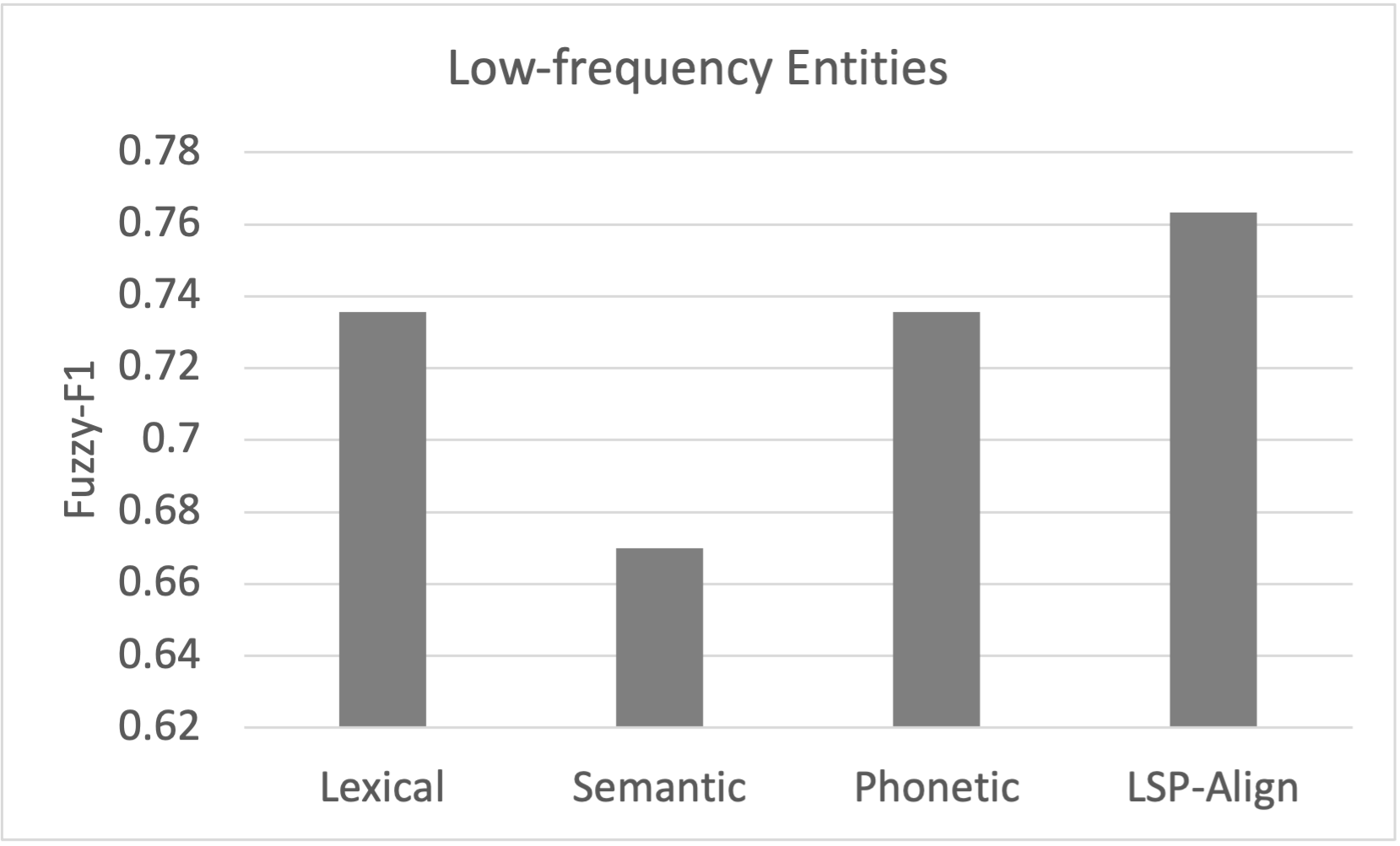}}}
\hspace{\fill}
 \subcaptionbox{mid-frequency \label{mid-freq}}{%
      \includegraphics[width=0.23\textwidth]{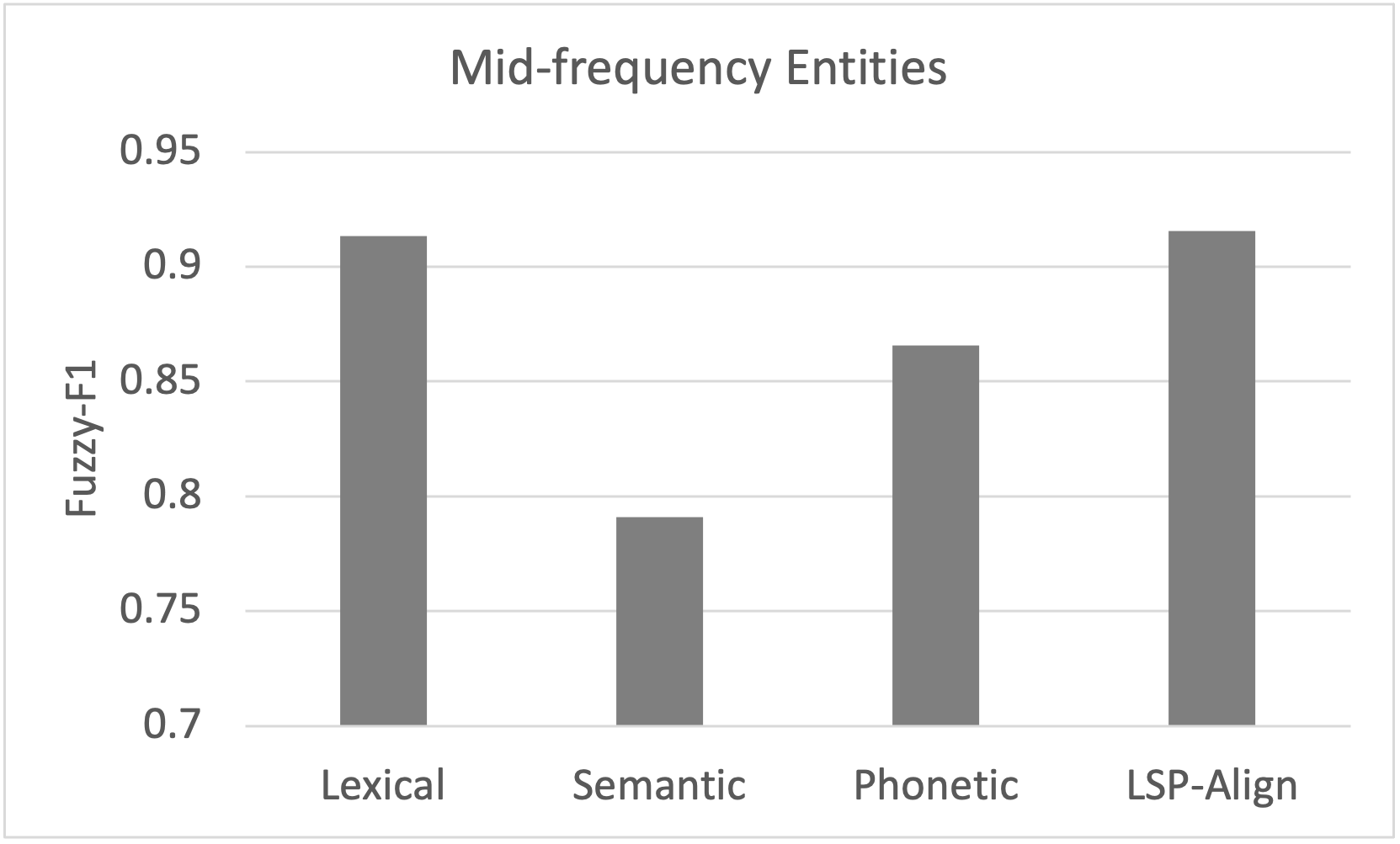}}
\hspace{\fill}
\center
  \subcaptionbox{high-frequency \label{high-freq}}{%
      \includegraphics[width=0.23\textwidth]{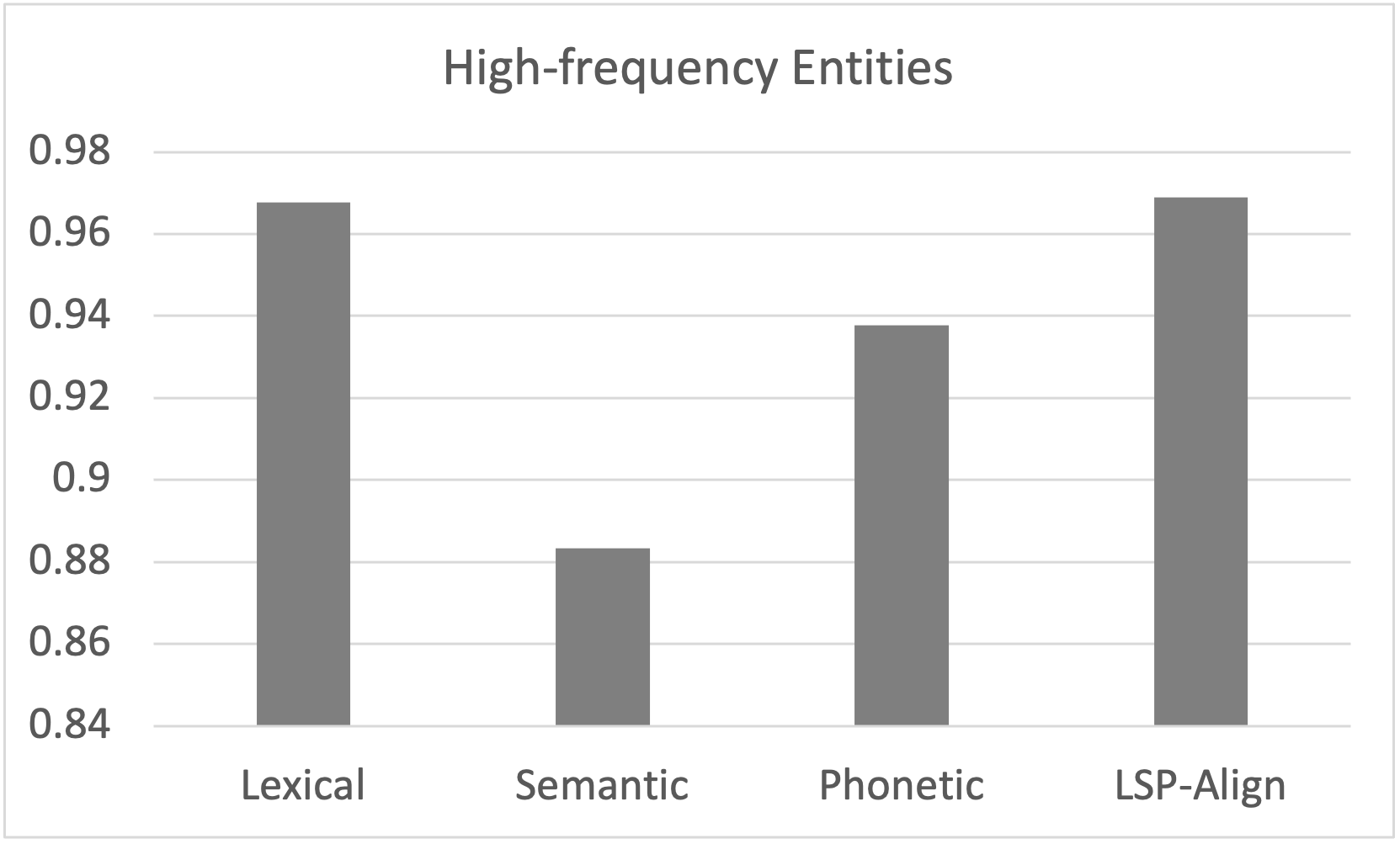}}\\
\caption{\label{per-freq} Fuzzy-F1 by entity-frequency}
\end{figure}
Figure~\ref{per-freq}, separates the evaluated entities by frequency in the web-data bitexts (low=0-3, mid=4-10, high=11+), and shows LSP-Align outperforming FastAlign when the entity is infrequent in the corpus. However, as entity frequency follows a long-tailed distribution, most entity mentions are infrequent. 

In Table~\ref{tab:full_data}, we evaluate the quality of our full XLEnt dataset. As a general trend, the quality of extracted entities is high-resource > mid-resource > low-resource. This is intuitive as there are more parallel sentences that are likely better aligned on a sentence-level yielding better word alignments.

\begin{figure}
\centering
\begin{small}
  \begin{tikzpicture}[scale=0.85]
    \begin{axis}[
    axis lines=middle,
    axis line style={->},
    x label style={at={(axis description cs:0.5,-0.1)},anchor=north},
    y label style={at={(axis description cs:-0.1,.5)},rotate=90,anchor=south},
    xlabel={Frequency Threshold},
    ylabel={Fuzzy-F1}]
      \addplot[
        mark=*,
        black
      ] plot coordinates {
        (1, 0.7881892854)
        (2, 0.8434470742)
        (3, 0.8561573018)
        (4, 0.86191075)
        (5, 0.8665196133)
      };\addlegendentry{High-resource}
      \addplot[
        mark=triangle*,
        black
      ] plot coordinates {
        (1, 0.5826901326)
        (2, 0.6582551869)
        (3, 0.6784071484)
        (4, 0.6910496546)
        (5, 0.6983993741)
      };\addlegendentry{Mid-resource}
      \addplot[
        mark=square*,
        black
      ] plot coordinates {
        (1, 0.4859112064)
        (2, 0.6056325987)
        (3, 0.6428137623)
        (4, 0.6571743102)
        (5, 0.671185547)
      };\addlegendentry{Low-resource}
    \end{axis}
  \end{tikzpicture}
  \end{small}
\caption{Fuzzy-F1 for high, mid, and low-resource languages
for different mined frequency thresholds.}
\label{fig:thresholds}
\end{figure}
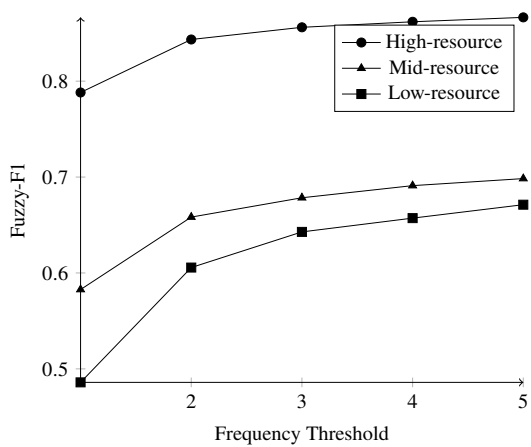

In Figure~\ref{fig:thresholds}, we show that filtering on a higher mined frequency improves the overall quality of the entity pairs (albeit yielding a smaller dictionary). This is also intuitive as the redundancy of an entity pair being mined multiple times in different sentence pairs signals it's likely a true translation. This suggests that tuning the frequency threshold can be a useful tool to control the quality of the resultant entity lexicon.
\begin{table*}[t]
\begin{minipage}[b]{0.28\linewidth}\centering
 \footnotesize
 
 \subcaptionbox{High-resource languages. \label{subtable:highresource_baseline}}{
 
\begin{tabular}{l@{\hspace{5pt}}l@{\hspace{8pt}}l@{\hspace{8pt}}l@{\hspace{8pt}}l@{\hspace{8pt}}}
\toprule
\textbf{Lang} & \textbf{Mined} & \textbf{P} & \textbf{R} & \textbf{F1} \\ \midrule
es  & 9.2M & 0.91 & 0.89 & 0.90 \\
fr  & 8.2M & 0.89 & 0.87 & 0.88 \\
ru  & 7.9M & 0.62 & 0.59 & 0.61 \\
ja  & 6.4M & 0.57 & 0.58 & 0.57 \\
zh  & 6.3M & 0.38 & 0.38 & 0.38 \\
nl  & 6.3M & 0.89 & 0.88 & 0.89 \\
it  & 6.1M & 0.90 & 0.88 & 0.89 \\
ar  & 5.8M & 0.78 & 0.77 & 0.77 \\
pt  & 5.8M & 0.90 & 0.89 & 0.90 \\
pl  & 5.5M & 0.85 & 0.85 & 0.85 \\
id  & 4.2M & 0.90 & 0.89 & 0.90 \\
de  & 4.1M & 0.87 & 0.85 & 0.86 \\
cs  & 3.9M & 0.86 & 0.88 & 0.87 \\
tr  & 3.8M & 0.87 & 0.84 & 0.86 \\
sv  & 3.7M & 0.91 & 0.91 & 0.91 \\
uk  & 3.7M & 0.75 & 0.68 & 0.72 \\
hu  & 3.6M & 0.87 & 0.82 & 0.85 \\
ro  & 3.3M & 0.90 & 0.89 & 0.90 \\
he  & 3.2M & 0.81 & 0.75 & 0.78 \\
da  & 3.0M & 0.92 & 0.92 & 0.92 \\
ca  & 3.0M & 0.90 & 0.88 & 0.89 \\
el  & 2.9M & 0.63 & 0.62 & 0.62 \\
ko  & 2.9M & 0.38 & 0.40 & 0.39 \\
hr  & 2.8M & 0.82 & 0.86 & 0.84 \\
fi  & 2.6M & 0.83 & 0.89 & 0.86 \\
sk  & 2.6M & 0.84 & 0.85 & 0.84 \\
bg  & 2.5M & 0.79 & 0.80 & 0.79 \\
eo  & 2.5M & 0.90 & 0.85 & 0.87 \\
no  & 2.2M & 0.89 & 0.90 & 0.89 \\
hi  & 2.0M & 0.73 & 0.74 & 0.73 \\
fa  & 1.9M & 0.59 & 0.60 & 0.59 \\
ms  & 1.8M & 0.85 & 0.79 & 0.82 \\
mk  & 1.8M & 0.83 & 0.86 & 0.84 \\
et  & 1.8M & 0.82 & 0.87 & 0.84 \\
gl  & 1.7M & 0.88 & 0.89 & 0.89 \\
lt  & 1.6M & 0.81 & 0.77 & 0.79 \\
bn  & 1.6M & 0.61 & 0.66 & 0.64 \\
sr  & 1.5M & 0.63 & 0.60 & 0.61 \\
\midrule 
\textbf{AVG} & 3.8M & 0.79	& 0.79 & 0.79 
\end{tabular}
}
\end{minipage}
    \hspace{0.75cm}
\begin{minipage}[b]{0.28\linewidth}\centering
 \footnotesize
 
 \subcaptionbox{Mid-resource languages. \label{subtable:midresource_baseline}}{
\begin{tabular}{l@{\hspace{5pt}}l@{\hspace{8pt}}l@{\hspace{8pt}}l@{\hspace{8pt}}l@{\hspace{8pt}}}
\toprule
\textbf{Lang} & \textbf{Mined} & \textbf{P} & \textbf{R} & \textbf{F1} \\ \midrule
sq  & 1.4M & 0.90 & 0.86 & 0.88 \\
lv  & 1.3M & 0.46 & 0.49 & 0.48 \\
th  & 1.2M & 0.20 & 0.31 & 0.24 \\
tl  & 1.1M & 0.81 & 0.78 & 0.79 \\
is  & 960K  & 0.86 & 0.82 & 0.84 \\
xh  & 960K  & 0.34 & 0.23 & 0.28 \\
sw  & 870K  & 0.79 & 0.84 & 0.82 \\
sl  & 860K  & 0.84 & 0.84 & 0.84 \\
eu  & 800K  & 0.80 & 0.76 & 0.78 \\
ur  & 746K  & 0.51 & 0.53 & 0.52 \\
ml  & 739K  & 0.37 & 0.45 & 0.41 \\
si  & 690K  & 0.66 & 0.67 & 0.67 \\
ast & 662K  & 0.58 & 0.62 & 0.60 \\
ta  & 644K  & 0.30 & 0.30 & 0.30 \\
be  & 582K  & 0.66 & 0.52 & 0.58 \\
mr  & 548K  & 0.46 & 0.53 & 0.49 \\
ha  & 437K  & 0.09 & 0.08 & 0.09 \\
mg  & 320K  & 0.72 & 0.79 & 0.76 \\
ne  & 319K  & 0.42 & 0.45 & 0.43 \\
lb  & 312K  & 0.46 & 0.49 & 0.47 \\
az  & 298K  & 0.67 & 0.68 & 0.67 \\
bs  & 267K  & 0.83 & 0.84 & 0.83 \\
ka  & 266K  & 0.48 & 0.47 & 0.47 \\
fy  & 266K  & 0.65 & 0.70 & 0.67 \\
jv  & 235K  & 0.59 & 0.64 & 0.62 \\
oc  & 234K  & 0.68 & 0.74 & 0.71 \\
af  & 234K  & 0.86 & 0.87 & 0.86 \\
cy  & 223K  & 0.84 & 0.81 & 0.82 \\
hy  & 216K  & 0.56 & 0.58 & 0.57 \\
ceb & 208K  & 0.32 & 0.39 & 0.35 \\
la  & 198K  & 0.57 & 0.59 & 0.58 \\
ga  & 171K  & 0.35 & 0.38 & 0.36 \\
kk  & 162K  & 0.60 & 0.59 & 0.59 \\
sd  & 148K  & 0.39 & 0.46 & 0.42 \\
te  & 147K  & 0.66 & 0.67 & 0.67 \\
su  & 144K  & 0.71 & 0.73 & 0.72 \\
yi  & 118K  & 0.11 & 0.18 & 0.14 \\
ht  & 115K  & 0.81 & 0.84 & 0.82 \\
\midrule 
\textbf{AVG} & 502K & 0.58 & 0.59 &	0.58 
\end{tabular}
}
\end{minipage}
    \hspace{0.75cm}
\begin{minipage}[b]{0.28\linewidth}\centering
 \footnotesize
 
 \subcaptionbox{Low-resource languages. \label{subtable:lowresource_baseline}}{
\begin{tabular}{l@{\hspace{5pt}}l@{\hspace{8pt}}l@{\hspace{8pt}}l@{\hspace{8pt}}l@{\hspace{8pt}}}
\toprule
\textbf{Lang} & \textbf{Mined} & \textbf{P} & \textbf{R} & \textbf{F1} \\ \midrule
br  & 109K  & 0.45 & 0.53 & 0.49 \\
mn  & 92K   & 0.63 & 0.62 & 0.62 \\
km  & 79K   & 0.37 & 0.46 & 0.41 \\
ilo & 79K   & 0.55 & 0.63 & 0.59 \\
am  & 72K   & 0.63 & 0.66 & 0.65 \\
so  & 70K   & 0.64 & 0.64 & 0.64 \\
ig  & 70K   & 0.67 & 0.69 & 0.68 \\
my  & 60K   & 0.43 & 0.35 & 0.38 \\
gd  & 58K   & 0.24 & 0.23 & 0.23 \\
nds & 57K   & 0.67 & 0.67 & 0.67 \\
ps  & 55K   & 0.64 & 0.60 & 0.62 \\
yo  & 51K   & 0.75 & 0.70 & 0.73 \\
fo  & 35K   & 0.76 & 0.79 & 0.77 \\
tt  & 33K   & 0.36 & 0.40 & 0.38 \\
ba  & 31K   & 0.44 & 0.43 & 0.44 \\
kn  & 31K   & 0.39 & 0.41 & 0.40 \\
gu  & 31K   & 0.52 & 0.52 & 0.52 \\
pa  & 29K   & 0.45 & 0.44 & 0.45 \\
lo  & 29K   & 0.40 & 0.37 & 0.38 \\
an  & 29K   & 0.74 & 0.76 & 0.75 \\
zu  & 28K   & 0.72 & 0.64 & 0.68 \\
bar & 25K   & 0.64 & 0.67 & 0.66 \\
or  & 23K   & 0.13 & 0.17 & 0.15 \\
arz & 22K   & 0.56 & 0.60 & 0.58 \\
wuu & 20K   & 0.06 & 0.32 & 0.11 \\
lmo & 20K   & 0.77 & 0.78 & 0.77 \\
io  & 18K   & 0.85 & 0.88 & 0.86 \\
tg  & 16K  & 0.34 & 0.36 & 0.35 \\
mwl & 14K  & 0.76 & 0.75 & 0.75 \\
wo  & 8K    & 0.29 & 0.28 & 0.29 \\
ff  & 7K    & 0.26 & 0.14 & 0.18 \\
ug  & 5K    & 0.15 & 0.18 & 0.16 \\
tn  & 5K    & 0.69 & 0.66 & 0.67 \\
as  & 3K    & 0.37 & 0.39 & 0.38 \\
ln  & 3K    & 0.27 & 0.28 & 0.27 \\
ss  & 2K    & 0.55 & 0.44 & 0.48 \\
om  & 1K    & 0.25 & 0.20 & 0.22 \\
lg  & 1K    & 0.22 & 0.19 & 0.20 \\
\midrule 
\textbf{AVG} & 35K & 0.49 & 0.50 & 0.49
\end{tabular}
}
\end{minipage}
\caption{Fuzzy metrics of extracted cross-lingual pairs evaluated against gold entity pairs. } 
\label{tab:full_data}
\end{table*}

\section{Conclusion}
We propose a technique that combines lexical alignment, semantic alignment, and phonetic alignment into a unified alignment model. We demonstrate this unified model better extracts cross-lingual entity pairs than any single alignment. Leveraging this model, we automatically curate a large, cross-lingual entity lexicon covering 120 languages paired with English which we freely release to the community. Accompanying this lexicon, we release a large multilingual collection of sentences tagged via named-entity projection. We hope these resources facilitate future multilingual NLP work such as multilingual NER, multilingual entity linking, and multilingual knowledge base construction.

\bibliography{main}
\bibliographystyle{acl_natbib}

\end{document}